\documentclass[sigconf,natbib=false]{acmart}

\AtBeginDocument{%
  }

\setcopyright{acmlicensed}
\copyrightyear{2024}
\acmYear{2024}
\setcopyright{acmlicensed}\acmConference[KDD '24]{Proceedings of the 30th ACM SIGKDD Conference on Knowledge Discovery and Data Mining}{August 25--29, 2024}{Barcelona, Spain}
\acmBooktitle{Proceedings of the 30th ACM SIGKDD Conference on Knowledge Discovery and Data Mining (KDD '24), August 25--29, 2024, Barcelona, Spain} \acmDOI{10.1145/3637528.3671455} \acmISBN{979-8-4007-0490-1/24/08}

\usepackage{bm}
\usepackage{multirow}
\usepackage[export]{adjustbox}

\usepackage{colortbl}
\definecolor{lightgray}{gray}{.9}
\definecolor{deepgray}{gray}{.8}

\usepackage{threeparttable}
\newcolumntype{I}{!{\vrule width 1pt}}
\makeatletter
\newcommand{\thickhline}{%
    \noalign {\ifnum 0=`}\fi \hrule height 1pt
    \futurelet \reserved@a \@xhline
}
\makeatother
\definecolor{mygray}{gray}{.9}
\usepackage{float}
\usepackage[ruled, vlined]{algorithm2e}
\usepackage{pifont}
\usepackage{ragged2e}
\usepackage{paralist}
\usepackage[toc,page]{appendix}
\usepackage{url}

\definecolor{mygray}{gray}{.9}
\definecolor{mygreen}{RGB}{93,173,85}
\definecolor{mywarning}{RGB}{233,144,61}
\definecolor{DarkRed}{RGB}{0,0,0}
\definecolor{azure}{rgb}{0.0, 0.5, 1.0}
\definecolor{gray}{rgb}{0.3, 0.3, 0.3}
\definecolor{DarkGreen}{RGB}{42,110,63}
\definecolor{linkblue}{RGB}{0, 0, 255} 
\newcommand{\hlg}[1]{\textcolor{mygreen}{#1}}
 
\newcommand{\pub}[1]{{\color{gray}{\footnotesize{[{#1}]}}}}

\newcommand{\tmark}{\ding{51}} 
\usepackage[capitalize]{cleveref}
\crefname{section}{Sec.}{Secs.}
\crefname{table}{Tab.}{Tabs.}
\crefname{section}{§}{§§}

\newcounter{bulletsec}
\renewcommand{\thebulletsec}{\arabic{bulletsec}}

\crefname{bulletsec}{§}{§§}
\Crefname{bulletsec}{§}{§§}

\makeatletter
\DeclareRobustCommand\onedot{\futurelet\@let@token\@onedot}
\def\@onedot{\ifx\@let@token.\else.\null\fi\xspace}

\def\eg{\emph{e.g}\onedot} 
\def\ie{\emph{i.e}\onedot}

\usepackage{amsmath}
\usepackage{mathtools}
\usepackage{amsthm}

\RequirePackage[
  datamodel=acmdatamodel,
  style=acmnumeric,
  sorting=none,
  ]{biblatex}

\begin{document}

\title{A Review of Graph Neural Networks in Epidemic Modeling}


\author{Zewen Liu}
\authornote{Equal Contribution}
\orcid{...}
\affiliation{
  \institution{Department of Computer Science}
  \institution{Emory University}
  \country{}
}
\email{zewen.liu@emory.edu}

\author{Guancheng Wan}
\authornotemark[1]
\orcid{...}
\affiliation{
  \institution{Department of Computer Science}   \institution{Emory University}
  \country{}
}
\email{gwan4@emory.edu}

\author{B. Aditya Prakash}
\orcid{...}
\affiliation{
  \institution{College of Computing}
  \institution{Georgia Institute of Technology}
  \country{}
}
\email{badityap@cc.gatech.edu}

\author{Max S. Y. Lau}
\orcid{...}
\affiliation{
  \institution{Department of Biostatistics and Bioinformatics}
  \institution{Emory University}
  \country{}
}
\email{msy.lau@emory.edu}

\author{Wei Jin}
\orcid{...}
\affiliation{
  \institution{Department of Computer Science}  \institution{Emory University}
  \country{}
}
\email{wei.jin@emory.edu}

\begin{abstract}

Since the onset of the COVID-19 pandemic, there has been a growing interest in studying epidemiological models. Traditional mechanistic models mathematically describe the transmission mechanisms of infectious diseases. However, they often suffer from limitations of oversimplified or fixed assumptions, which could cause sub-optimal predictive power and inefficiency in capturing complex relation information. Consequently, Graph Neural Networks (GNNs) have emerged as a progressively popular tool in epidemic research. In this paper, we endeavor to furnish a comprehensive review of GNNs in epidemic tasks and highlight potential future directions. To accomplish this objective, we introduce hierarchical taxonomies for both epidemic tasks and methodologies, offering a trajectory of development within this domain. For epidemic tasks, we establish a taxonomy akin to those typically employed within the epidemic domain. For methodology, we categorize existing work into \textit{Neural Models} and \textit{Hybrid Models}. Following this, we perform an exhaustive and systematic examination of the methodologies, encompassing both the tasks and their technical details. Furthermore, we discuss the limitations of existing methods from diverse perspectives and systematically propose future research directions. This survey aims to bridge literature gaps and promote the progression of this promising field, with a list of relevant papers at \textcolor{linkblue}{\url{https://github.com/Emory-Melody/awesome-epidemic-modeling-papers}}. We hope that it will facilitate synergies between the communities of  GNNs and epidemiology, and contribute to their collective progress.


\end{abstract}






\maketitle







\section{Introduction}

\begin{figure*}[t]
\centering
\includegraphics[width=\linewidth]{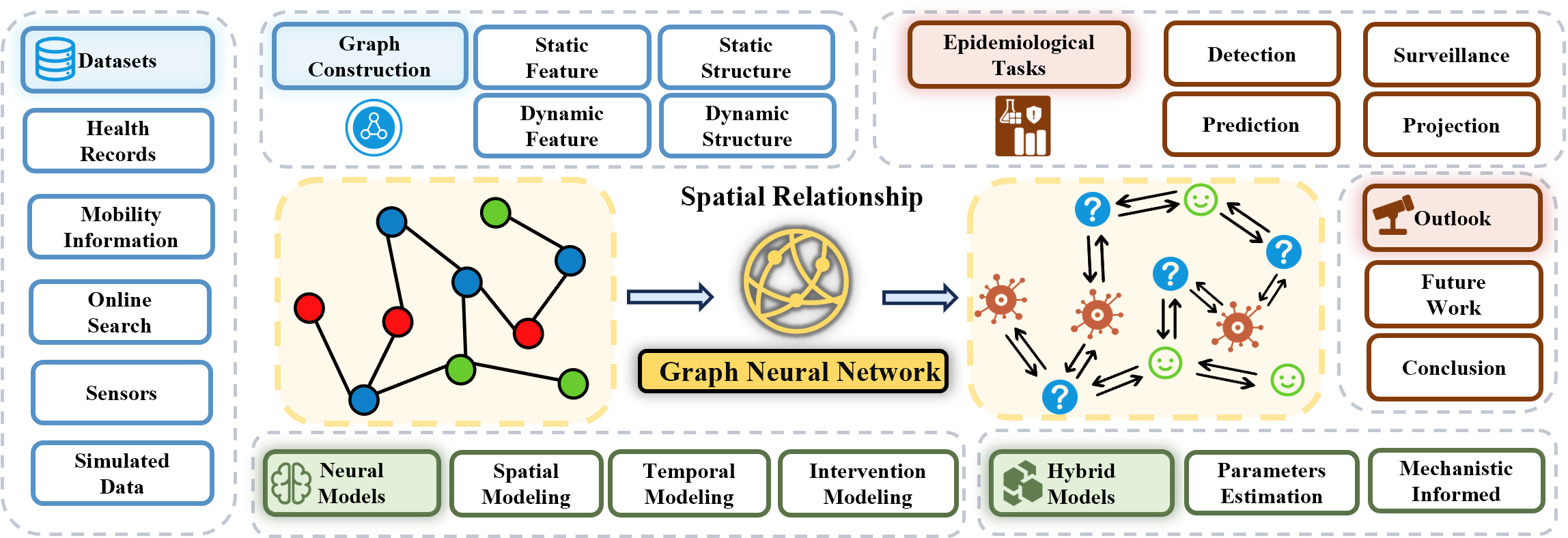}
\vspace{-10pt}
\caption{
\textbf{Overview of the survey}. Best viewed in color.
}  
\label{fig:framework}
\vspace{-10pt}
\end{figure*}


Epidemiology has long been a critical field, with its origins tracing back to ancient societies that observed patterns of disease spread~\cite{bramanti2012ancient, bruce-chwatt1977plagues}. Although the conceptualization of epidemiology has evolved over time, the terms \emph{health} and \emph{control} have been predominantly associated with it since 1978~\cite{frerot2018epidemiology}. Currently, the World Health Organization (WHO) describes epidemiology as the investigation into the distribution and determinants of health-related states or events, encompassing a broad spectrum of issues including disease transmission, vaccination efforts~\cite{fine2015another, terris1993society}, cancer and diabetes treatment, etc. This definition underscores the field's emphasis on controlling health-related issues and making informed decisions. A pertinent illustration of this is the COVID-19 pandemic, which rapidly infected millions worldwide, placing immense strain on the production and distribution of medical resources~\cite{cm2020does, bergrath2022impact}. Decision-making processes like allocation of resources, are crucial in mitigating the impact of diseases and saving lives, highlighting the importance of advancements in epidemiology~\cite{emanuel2020fair}.


To address a range of health-related challenges, there is an indispensable need for epidemic modeling, and researchers have devised various mechanistic models~\cite{funk2018realtime, kondratyev2013forecasting}. These models, grounded in mathematical formulations, simulate the dissemination of infectious diseases by incorporating biological and behavioral underpinnings. By considering factors such as population, they yield insights into patterns of disease transmission and the efficacy of intervention strategies, thereby playing a pivotal role in shaping public health policies~\cite{louz2010emergence, mikolajczyk2009influenza}. However, these knowledge-driven methods often depend on oversimplified or fixed assumptions that can lead to biases in modeling. Consequently, this compromises both the accuracy of predictions and their ability to generalize across different contexts.




To overcome the limitations of mechanistic models, there is an emerging trend to adopt data-driven approaches in epidemic forecasting tasks,  with a particular emphasis on machine learning and deep learning models~\cite{shorten2021deep}. Specifically, Convolutional Neural Networks (CNNs) and Recurrent Neural Networks (RNNs) have demonstrated great success in various epidemiological predictive tasks, including forecasting daily new case counts, estimating virus reproduction and doubling times, and determining disease-related factors~\cite{wu2018deep, saleem2022machine, baldo2021deep}. Despite their effectiveness in these tasks, these models often fall short in incorporating relational information from critical epidemiological data sources such as human mobility, geographic connections, and contact tracing. This deficiency restricts their utility in broader epidemiological applications. 

Recently, the advances in Graph Neural Networks (GNNs)~\cite{SurveyonGNN_TNNLS20,GCN_ICLR17,GAT_arxiv17,GATv2_ICLR22} have set the stage for overcoming the aforementioned challenges in epidemic modeling. Specifically, GNNs stand out for their ability to aggregate diverse information through a message-passing mechanism, making them particularly suited for capturing relational dynamics within graphs~\cite{liu2023uplifting, maskey2022generalization}. Thus, by representing interactions between entities as graphs, researchers can leverage GNNs to harness relational data effectively and facilitate epidemiology tasks~\cite{deng2020colagnn,wang2022causalgnn}. 
For instance, GNNs are often utilized to model spatial interactions~\cite{gao2021stan} and other complex interactions~\cite{caoMepoGNNMetapopulationEpidemic2022}, enhancing the analysis of graph data and yielding more precise predictions. In addition, GNNs bring a certain level of interpretability by quantifying the influence of individual nodes (or entities) on final prediction~\cite{feng2023contact}. Moreover, the flexible design of GNNs facilitates their integration with traditional mechanistic and probabilistic models to leverage expert knowledge and offer measures of uncertainty. As a result, GNNs have found extensive applications in various tasks within the field including infection prediction ~\cite{liu2023epidemiologyaware} , outbreak source detection~\cite{ru2023inferring}, intervention modeling~\cite{song2020reinforced}, etc., facilitating advancements in epidemiology research.

Considering the critical role of epidemic modeling and the widespread adoption of GNNs in this area, a systematic review of these algorithms is essential for advancing our understanding of the field. We seek to bridge this knowledge gap by providing a thorough overview and categorization of how GNNs are applied in epidemiological studies. Our goal extends beyond merely highlighting current research directions; we aim to uncover new directions for future exploration and enrich both the GNN (or graph machine learning) and epidemiology communities. 



\begin{table}[t]

\caption{{
\textbf{Overview of related surveys}.
}
}
\label{tab:ourscomputation}
\vspace{-10pt}
\centering
{
\resizebox{\columnwidth}{!}{
\setlength\tabcolsep{2.5pt}
\renewcommand\arraystretch{1.2}
\begin{tabular}{rIcccc}
\hline\thickhline
 Work & Tasks Taxonomy & GNNs & Mechanistic Model & Future Work\\
\hline\hline 
\pub{ArXiv'21} \cite{baldo2021deep} &  & \hlg{\tmark{}} & \hlg{\tmark{}} & \\
\pub{JBD'21}\cite{shorten2021deep} & \hlg{\tmark{}}  & \hlg{\tmark{}} & \hlg{\tmark{}} & \\ 
\pub{RBME'22}\cite{clement2022survey} &  &  & \hlg{\tmark{}} & \hlg{\tmark{}}\\ 
\pub{NC'22} \cite{kamalov2022deep} &  & \hlg{\tmark{}} & \hlg{\tmark{}} &  \\ 
\pub{RBE'22} \cite{dinesh10significance} & \hlg{\tmark{}} &  &  &  \\ 
\pub{IJERPH'22} \cite{saleem2022machine} & \hlg{\tmark{}} &  & \hlg{\tmark{}} &  \\ 
\pub{ArXiv'22} \cite{Data-centric_arxiv22} & \hlg{\tmark{}} & \hlg{\tmark{}} & \hlg{\tmark{}} &  \hlg{\tmark{}}\\ 
\hline \hline
Ours & \hlg{\tmark{}} & \hlg{\tmark{}} &  \hlg{\tmark{}} & \hlg{\tmark{}}\\
\end{tabular}}}
\label{tab: Connect_Survey}
\vspace{-10pt}
\end{table}

\vspace{3mm}
\noindent
\textbf{Contributions.} This paper presents a comprehensive review of GNNs in epidemic modeling. We focus on task categorization, summarizing the latest methodologies, and outlining future directions. We aspire for this work to become a valuable asset for researchers keen on exploring this interdisciplinary research direction. Our contributions are summarized as follows:

\begin{compactenum}[(1)]
  \item Our work offers a comprehensive and pioneering review of GNNs in the context of epidemic modeling. It encompasses a detailed categorization of various tasks, data resources, and graph construction techniques in the field, as elaborated from Sections~\ref{sec: Epide_Tasks} to \ref{sec: Graph_Construct}.
  \item We offer an in-depth classification of existing methodologies, complemented by a meticulous review in Section~\ref{sec: Method}.
  \item We point out current methods' limitations and provide prospective directions in Section \ref{sec: Future_Work}, thereby facilitating the ongoing progression of the community.
\end{compactenum}

\vspace{3mm}
\noindent
\textbf{Connections to Existing Surveys.} In contrast to previous surveys that explore the intersection of epidemiology and deep learning models, our paper offers a detailed overview specifically of GNNs in epidemic modeling. While preceding surveys predominantly concentrate on predicting COVID-19 outcomes and often overlook the inclusion of GNN-based methodologies \cite{shorten2021deep, clement2022survey, dinesh10significance, saleem2022machine}, a few studies have indeed integrated GNNs \cite{kamalov2022deep, baldo2021deep}. However, the scope of such works remains relatively narrow, predominantly confined to virus transmission tasks. Certain studies concentrate exclusively on a singular virus~\cite{saleem2022machine} or are dedicated to a specific task like forecasting~\cite{Data-centric_arxiv22}. Distinctively, our research is tailored towards GNN-based approaches, covering a broader spectrum of epidemic modeling tasks. Furthermore, our survey presents the latest review of GNN applications in epidemic modeling, offering deeper insights when compared to existing literature. The comparison can be found in Table~\ref{tab: Connect_Survey}. Figure~\ref{fig:framework} shows the structure of this survey.

\section{preliminaries and definitions}
\subsection{Learning on Graph Data}
In this paper, we define the graph data as \( G = (V, \mathcal{E})\),
with \( V \) representing the node set comprising \( |V|= N\) nodes. The edge set \( \mathcal{E} \subseteq V \times V \) represents the connections between nodes. The feature matrix \( \textbf{X} = \{\textbf{x}_1, \textbf{x}_2, \ldots, \textbf{x}_N\}^\top \in \mathbb{R}^{N \times D}\) includes feature vectors \( \textbf{x}_i \) associated with node \( v_i \), where $D$ denotes the feature dimension. The adjacency matrix of $G$, denoted by $\textbf{A} \in \mathbb{R}^{N \times N}$, sets $\textbf{A}_{i j}=1$ for any existing edge $e_{i,j} \in \mathcal{E}$ and $\textbf{A}_{i j}=0$ otherwise. The normalized adjacency matrix is given by \( \hat{\textbf{A}} = \textbf{D}^{-1/2}\textbf{A}\textbf{D}^{-1/2} \). The degree matrix \(\textbf{D} \), being a diagonal matrix, is characterized by \( \textbf{D}_{i,i} = \sum_j \textbf{A}_{i,j} \).

In the domain of graph learning, the node-level task stands out as a significant area of focus. The objective of this task is to forecast the properties (\ie{}, numerical value or probability) or class of the individual nodes. This process entails training a neural network model that utilizes a subset of nodes with known properties, denoted as $\mathcal{V}_L$, to infer the properties of other unknown nodes. The essence of this training is encapsulated by optimizing the function:
\begin{equation}
\label{eq: node_train}
\min_{\theta} \mathcal{L} (f_{\theta}(G)) = \sum_{v_i \in \mathcal{V}_L} \ell (f_{\theta}(\textbf{X}, \textbf{A})_i; y_i),
\end{equation}
Here, the function $f_{\theta}(\textbf{X}, \textbf{A})$ aims to forecast the property for each node, with $y_i$ representing the actual state of node $v_i$. The discrepancy between the predicted and true properties is quantified using a loss function $\ell(\cdot, \cdot)$, such as RMSE (Root Mean Square Error).

\subsection{Graph Neural Networks}
Over recent years, GNNs have garnered increasing interest and have been deployed across diverse fields, including bioinformatics, material science, chemistry, and neuroscience~\cite{reiser2022graph, wen2022graph, wieder2020compact, bessadok2022graph}. Among them, Graph Convolutional Networks (GCN)~\cite{GCN_ICLR17} and Graph Attention Networks (GAT)~\cite{GAT_arxiv17, GATv2_ICLR22}, have advanced the frontier of research on graph-structured data with their sophisticated and effective designs~\cite{SurveyonGNN_TNNLS20, TrustworthyGNN_survey_arxiv22, deep_graph_clustering_survey_arxiv}. Typically, GNNs aim to learn graph representations, including node embeddings $\textbf{h}_i \in \mathbb{R}^{d}$, by utilizing both the structural and feature information of a graph $G$. The process within a GNN involves two key operations: message passing and aggregation of neighborhood information. This involves each node in the graph repeatedly collecting and integrating information from its neighbors as well as its own attributes to enhance its representation. The operation of a GNN over $L$ layers can be described by the following expression:
\begin{equation}
\textbf{h}^{(l+1)}_i = \sigma (\textbf{h}^{(l)}_i, \mathit{AGG}({\textbf{h}^{(l)}_j; j \in \textbf{A}_i})), \forall l\in [L],
\end{equation}
where $\textbf{h}^{(l)}_i$ is the representation of node $v_i$ at layer $l$, with $\textbf{h}^{(0)}_i= \textbf{x}_i$ being the initial node features. Here, $\textbf{A}_i$ represents the set of neighbors for node $v_i$, $\mathit{AGG(\cdot)}$ denotes a variant-specific aggregation function, and $\sigma$ represents an activation function. Following the completion of $L$ layers of message passing, the resultant node embedding $h_i$ is passed through a projection function $F(\textbf{h}_i)$ to produce the output prediction $\hat{y}_i$.

\subsection{Mechanistic Models}

Empirical models~\cite{aleta2020modelling, chang2020modelling} in epidemic forecasting utilize historical data to discern patterns and forecast the future spread of diseases. In contrast, mechanistic models~\cite{jiang2021countrywide, yang2023epimob} provide a detailed framework that explores the biological and social complexities underlying the transmission of infectious diseases, thus exceeding the reliance on historical data inherent to empirical models. Among mechanistic approaches, classic compartmental models~\cite{1927contribution} (\eg{}, SIR) are particularly notable. These models adeptly simplify the intricate dynamics of disease transmission into digestible components. This simplification facilitates a clearer understanding of infection spread, serving as a valuable tool for both researchers and policymakers.

\vskip 0.3em
\noindent\textbf{SIR Compartmental Model}.
In the domain of epidemiology~\cite{danon2011networks, caals2017ethics}, it is widely hypothesized that the rate at which networks evolve is significantly slower compared to the propagation speed of diseases. This fundamental assumption underpins the adoption of a SIR model~\cite{1927contribution, dehning2020inferring}, which is instrumental in accurately capturing the dynamics of epidemic spread. The SIR model categorizes the population into three distinct groups based on their disease status: susceptible (S) to infection, currently infectious (I), and recovered (R), with the latter group being immune to both contraction and transmission of the disease. The SIR model, formulated using ordinary differential equations (ODEs)~\cite{grassly2008mathematical}, are as follows:
\begin{equation}
\begin{gathered}
\label{eq: SIR}
\setlength\abovedisplayskip{5pt} \setlength\belowdisplayskip{5pt}
\frac{dS(t)}{dt} = -\beta \frac{S(t)I(t)}{N}, \\
\frac{dI(t)}{dt} = \beta \frac{S(t)I(t)}{N} - \gamma I(t), \quad
\frac{dR(t)}{dt} = \gamma I(t).
\end{gathered}
\end{equation}
These equations distribute the total population \(N\) across the aforementioned categories, with the transitions between states regulated by two pivotal parameters: the transmission rate \(\beta\) ($S \to I$) and the recovery rate $\gamma$ ($I \to R$). The model posits a quadratic relationship for disease transmission via interactions between susceptible and infectious individuals (\(\beta S(t)I(t)\)), alongside a linear recovery mechanism (\(\gamma I(t)\)). By fine-tuning the parameters of the SIR model, it is possible to compute the basic reproduction number \(R_0 = \beta/\gamma\), serving as a metric for the disease transmission potentials~\cite{hanDevilLandscapesInferring2023, shah2020finding}.

Furthermore, utilizing the SIR model within the context of graph-based structures leads to the development of what is referred to as the \textbf{Network SIR} model~\cite{sha2021source, brede2012networks, balcan2009multiscale, venkatramanan2017spatiotemporal}. Within this framework, an infectious node \(j\) has the potential to transmit the infection to another node \(i\), provided that \(i\) is susceptible and located adjacently to \(j\) (\ie{}, \(i \in A(j)\) ). The probabilities of the node \(i\) being in a susceptible \(S_i(t)\), infected \(I_i(t)\), or recovered \(R_i(t)\) state at any given time \(t\), they are recalculated as follows:
\begin{equation}
\begin{gathered}
\label{eq: network_SIR}
\setlength\abovedisplayskip{5pt} \setlength\belowdisplayskip{5pt}
\frac{dS_i(t)}{dt} = -\beta \sum_{j} A_{ij}S_i(t)I_j(t), \\
\frac{dI_i(t)}{dt} = \beta \sum_{j} A_{ij}S_i(t)I_j(t) - \gamma I_i(t), \quad
\frac{dR_i(t)}{dt} = \gamma I_i(t).
\end{gathered}
\end{equation}
Considering the initial phase where \(S_i(t)\) is nearly one, the model facilitates the calculation of the basic reproduction number \(R_0\) as \(R_0 = \beta \lambda_1 / \gamma\), with \(\lambda_1\) representing the leading eigenvalue of the adjacency matrix \(A\).

\vskip 0.3em
\noindent\textbf{SIR Variants.} 
While the SIR model provides a powerful framework for analyzing disease dynamics, its simplicity can sometimes neglect critical factors such as incubation periods, non-permanent immunity, and heterogeneous population interactions. This limitation has spurred the development of SIR variants, which offer a more comprehensive and nuanced understanding of the spread and control of infectious diseases. We briefly outline some of the most commonly used variants:
\noindent \textbf{i)} SEIR \cite{dixon2018seds}: The SEIR model extends the basic SIR framework by incorporating an 'Exposed' (E) compartment. This compartment represents individuals who have been exposed to an infectious disease but are not yet infectious themselves~\cite{1927contribution, vandendriessche2017reproduction}. The model details the transition through the stages according to the sequence: \(S \rightarrow E \rightarrow I \rightarrow R\).
\noindent \textbf{ii)} SIRD: Enhancing the traditional SIR model, the SIRD variant includes a 'Dead' (D) compartment, thus adapting the progression to: \(S \rightarrow I \rightarrow R \rightarrow D\). This modification accounts for individuals who succumb to the disease, providing a more accurate depiction of its mortality impact~\cite{tomyEstimatingStateEpidemics2022, wangCausalGNNCausalBasedGraph2022, lolipiccolomini2020monitoring}.

\noindent\textbf{Bridging the Gap.} To bridge the gaps between mechanistic models and neural models like GNNs, a Python library called \textbf{EpiLearn} ~\cite{liu2024epilearn} is also developed at \url{https://github.com/Emory-Melody/EpiLearn}. This library serves as a toolkit for epidemic modeling and analysis, empowering data mining on epidemiology data with machine learning models.



\begin{table}[t]
\caption{
\small{\textbf{A brief description of epidemic tasks we categorized}.}
}
\label{tab: tasks}
\vspace{-10pt}
\centering
{
\resizebox{\linewidth}{!}{
\setlength\tabcolsep{10pt}
\renewcommand\arraystretch{1.2}

\begin{tabular}{@{}c|c|c@{}}
\hline
\thickhline

\textbf{Tasks}  & \textbf{Time Interval} & \textbf{Objective} \\ 
\hline \hline
\textbf{Detection} & History-Present & Incident Back-tracing \\
\textbf{Surveillance} & Present & Event Monitoring \\
\textbf{Prediction} & Future & Future Incident Prediction \\
\textbf{Projection} & Future &  Change Simulation and Prediction\\

\hline
\end{tabular}
}
}
\vspace{-15pt}
\end{table}

\begin{table*}[t]
\caption{
\small{\textbf{Summary of epidemiological tasks and representative GNN-based methods}.}
}
\label{tab: summary}
\vspace{-10pt}
\centering
{
\resizebox{0.9\linewidth}{!}{
\setlength\tabcolsep{10pt}
\renewcommand\arraystretch{1.2}

\begin{tabular}{@{}c|c|c|c|c@{}} 
\hline
\thickhline

\textbf{Task}                          & \textbf{Paper} & \textbf{Methodology} & \textbf{Hybrid} & \textbf{Graph Construction} \\ 
\hline \hline
\multirow{1}{*}{\textbf{Detection}}    
                              &   SD-STGCN ~\cite{sha2021source}    & GAT + GRU + SEIR & \hlg{\tmark{}} & Spatial-Temporal Graph; Static Graph Structure    \\

\cline{1-5}
\multirow{2}{*}{\textbf{Surveillance}} 
                              &   WDCIP ~\cite{wang2023wdcip}     &  GAE &  & \multirow{1}{*}{Spatial Graph; Static Graph Structure}          \\
                              &   GraphDNA~\cite{yang2022dynamic} & GCN + LSTM & & \multirow{1}{*}{Spatial-Temporal Graph; Dynamic Graph Structure} \\
   
\cline{1-5}
\multirow{4}{*}{\textbf{Projection}}   
                              &   MMCA-GNNA~\cite{jhun2021effective}    & GNN + SIR + RL & \hlg{\tmark{}} & \multirow{2}{*}{Spatial-Temporal Graph; Static Graph Structure}   \\ 
                              &   DURLECA ~\cite{song2020reinforced}    &  GNN + RL  &    & \\  \cline{5-5}
                              
                              &   IDRLECA ~\cite{fengContactTracingEpidemic2023}    & GNN + RL &  & \multirow{1}{*}{Spatial-Temporal Graph; Dynamic Graph Structure}               \\
                              
\cline{1-5}
\multirow{19}{*}{\textbf{Prediction}}
                              &    DGDI ~\cite{liu2023human}   & GCN + Self-Attention &  & \multirow{1}{*}{Spatial Graph; Static Graph Structure}       \\ \cline{5-5}
                              &    DVGSN ~\cite{zhang2023predicting}   & GNN &  &Temporal-Only Graph; Static Graph Structure               \\ \cline{5-5}
                              
                              &   STAN ~\cite{gao2021stan} & GAT + GRU &  & \multirow{8}{*}{Spatial-Temporal Graph; Static Graph Structure}             \\
                              &    MSDNet ~\cite{tang2023enhancing}   &  GAT + GRU + SIS  &  \hlg{\tmark{}} &            \\
                              &    SMPNN ~\cite{lin2023grapha}   &  MPNN + Autoregression &    &           \\    
                              &    ATMGNN ~\cite{nguyen2023predicting}   & MPNN/MGNN + LSTM/Transformer &     &          \\
                              &    DASTGN ~\cite{puDynamicAdaptiveSpatio2023}   &   GNN + Attention + GRU   &  &       \\
                              &    MSGNN ~\cite{qiu2023msgnna}   &  GCN + N-Beats  &      &         \\
                              &    STEP ~\cite{yuSpatiotemporalGraphLearning2023}   &  GCN + Attention + GRU  &   &    \\
                              &    GSRNN ~\cite{li2019study}   & GNN + RNN &  & \\   \cline{5-5}

                              &   Mepo GNN~\cite{cao2023metapopulation, caoMepoGNNMetapopulationEpidemic2022} & (TCN + GCN) + Modified SIR & \hlg{\tmark{}} & \multirow{10}{*}{Spatial-Temporal Graph; Dynamic Graph Structure}  \\
                              &   Epi-Cola-GNN ~\cite{liu2023epidemiologyaware} &  Cola-GNN + Modified SIR  &  \hlg{\tmark{}} &  \\
                              &   HiSTGNN ~\cite{ma2022hierarchical} &  Hierarchical GNN + Transformer  &    \\
                              &    CausalGNN ~\cite{wang2022causalgnn} & GNN + SIRD & \hlg{\tmark{}} &    \\
                              &    ATGCN ~\cite{wang2022adaptively}   &   GNN + LSTM   &   &  \\
                              &    HierST ~\cite{zheng2021hierst}   & GNN + LSTM  &   &   \\
                              &    RESEAT ~\cite{moon2023reseat}   & GNN + Self-Attention &  &  \\
                              &    SAIFlu-Net ~\cite{jungSelfAttentionBasedDeepLearning2022}   & GNN + LSTM &   &   \\
                              &    Epi-GNN ~\cite{xieEpiGNNExploringSpatial2022}   & GCN + Attention + RNN &  &  \\
                              &    Cola-GNN ~\cite{deng2020colagnn}   & GCN + Attention + RNN &    &   \\

\cline{1-4}
\hline
\end{tabular}}}
\vspace{-10pt}
\end{table*}

\section{Taxonomies}
In this section, we provide taxonomies for GNNs in epidemic modeling. These methods can be categorized into different types based on their epidemiological tasks, datasets, graph construction, and methodological distinctions. A comprehensive categorization is shown in Appendix \ref{app: 1} and due to page limitation, we provide part of it in Table~\ref{tab: summary}.

\subsection{Epidemiological Tasks}
\label{sec: Epide_Tasks}
For epidemiological tasks, we provide a taxonomy from the perspective of epidemiologists and categorize the work we investigated into four categories based on researchers' goals: \textbf{Detection}, \textbf{Surveillance}, \textbf{Prediction}, and \textbf{Projection}. A brief comparison of these tasks is shown in Table~\ref{tab: tasks}; the detailed explanations and definitions are introduced as follows:

\subsubsection{Detection}
The goal of detection tasks is to identify health states, disease spread, or other related incidents that \underline{happened at a} \underline{specific time}. In this survey, we incorporate two different detection tasks from the view of graph data: \textit{source detection} and \textit{transmission detection}. To formulate a mathematical definition, the temporal network, which consists of sequenced graphs from different time points, is represented as $G = \{G_0, G_1, \ldots, G_T\}$. Within a graph $G_t$, the states of nodes and edges are represented by $S_t^V$ and $S_t^\mathcal{E}$ respectively. Then, the detection task can be expressed as predicting $S_t^V$ or $S_t^\mathcal{E}$ given graph $G_T$ and time point $t$. 

For example, finding patient-zero ~\cite{ruInferringPatientZero2023, sha2021source, shah2020finding}, as a source detection task, is important for identifying the source of disease outbreaks and aims to find a set of nodes $V$ on graph $G_0$. In this setting, the problem can also be seen as identifying the state of each node at the initial time point, which is $S_0^V$.

\subsubsection{Surveillance}
Surveillance tasks aim at providing timely and accurate information to support decision-making and disease prevention. Since a prompt response is needed, \underline{real-time processing} \underline{ability} has been the most important requirement during modeling. Here, we provide a formal definition: given a temporal graph $G = \{G_0, G_1, \ldots, G_T\}$, the goal is to identify a target statistic $\textbf{y}$ on graph $G_T$ at the present moment or in the short term.

To illustrate, tasks like detecting infected individuals promptly \cite{songCOVID19InfectionInference2023} and estimating infection risks in different locations in real-time~\cite{wang2023wdcip, yang2022dynamic, gouarebDetectionPatientsRisk2023, hanDevilLandscapesInferring2023} can be seen as surveillance tasks, as they their prediction targets lie in present or near future. 

\subsubsection{Prediction}
Similar to surveillance tasks, prediction tasks also aim to forecast epidemic events using historical data. However, unlike surveillance tasks, prediction tasks typically involve \underline{longer time spans} and \underline{do not require real-time processing}. Therefore, prediction tasks are more interested in predicting the target at the longer time ahead like $T+1$ instead of at time $T$. Due to the large amount of work, we further classify prediction tasks into two categories based on the type of prediction target: 

\begin{compactenum}[(1)]
\item \textit{Incidence Prediction.} The target of incidence prediction is to provide quantitative results. In epidemic forecasting, incidences can include the number of infections or deaths in the future ~\cite{sijiraniSpatioTemporalPredictionEpidemiology2023, yu2023spatiotemporal, nguyenPredictingCOVID19Pandemic2023, croftForecastingInfectionsSpatiotemporal2023, puDynamicAdaptiveSpatio2023, qiu2023msgnna, moonRESEATRecurrentSelfAttention2023, caoMetapopulationGraphNeural2023, tangEnhancingSpatialSpread2023}, influenza activity level ~\cite{liuEpidemiologyawareDeepLearning2023}, Influenza-Like Illness (ILI) rates ~\cite{zhangPredictingInfluenzaPandemicawareness2023}, vaccine hesitancy ~\cite{moonGraphBasedDeep2023}, etc. The prediction of these incidences is important to decision-making, proactive public health planning, and the effective management of infectious diseases and other health challenges. 

\item \textit{Trend Prediction.} Different from incidence prediction task, which focuses on quantitative targets, the target of trend prediction tasks is to identify a higher-level epidemic spreading pattern. For transmission among locations ~\cite{liu2023human}, prediction of infection trend can be described as an information retrieving problem and the goal is to predict the next region to be infected given a historic spreading route. For transmission among individuals or groups, the goal usually includes identifying transmission dynamics in emerging high-risk groups ~\cite{sunDeepDynaForecastPhylogeneticinformedGraph2023}.
\end{compactenum}

\subsubsection{Projection}
In epidemic forecasting, projection tasks are similar to prediction, but with \underline{an additional intention to understand} \underline{epidemic outcomes}. These tasks usually require models with the ability to incorporate changes during the evolving of epidemics, such as external interventions and changing of initial states. Most of the projection tasks we collected involve finding the \textit{optimal interventions} or \textit{maximizing influence} to achieve targets like curbing the spread of diseases. Influence maximization ~\cite{kempe2003maximizing} aims to identify a subset of nodes so that the influence spreads most effectively across the graph, and there have been several early studies in epidemiological tasks, e.g., node importance ranking ~\cite{bucur2020beyond, holme2017three}. 

In this paper, we extend the traditional setting of influence maximization and combine it with intervention strategy tasks to form a more general definition as follows: Given a temporal graph $G = \{V(t), \mathcal{E}(t)\}$, the states of nodes $\in V$ and edges $\in \mathcal{E}$ are influenced by strategies defined as $P_v(t)$ and $P_\mathcal{E}(t)$, which represents strategies on nodes and edges respectively. The goal of the task is to find optimal strategies so that the target is maximized or minimized.

For traditional influence maximization tasks and vaccine strategy tasks ~\cite{jhun2021effective}, which aim to vaccinate the optimal set of nodes to minimize epidemic damage, strategies are limited to nodes at the starting time point, which is $P_v(0)$. For interventions throughout the period, strategies can include applying quarantine level to nodes at each step ~\cite{fengContactTracingEpidemic2023}, which denotes $P_v(t)$ or restricting mobility on edges ~\cite{meirom2021controlling, song2020reinforced}, which denotes $P_\mathcal{E}(t)$.

\subsubsection{Perspectives from Data Scientists.}

The above general taxonomy for epidemiological tasks comes from the perspective of epidemiologists. However, it is also feasible to categorize these works from the perspective of data scientists, who focus more on the computational pipeline. Here we provide a further taxonomy from the perspective of model \textbf{inputs} and \textbf{outputs}:

\begin{compactenum}[(1)]
    \item For inputs of all models, they all consist mainly of two parts: node features and the graph structure. Based on the temporality of nodes, we can further categorize these work into \textbf{spatial-only} tasks, \textbf{temporal-only} ~~\cite{zhangPredictingInfluenzaPandemicawareness2023}, and \textbf{spatial-temporal} tasks. In addition, based on the temporality and learnability of graph structure, we can also use \textbf{static} or \textbf{dynamic} features to distinguish these works. A detailed introduction is presented below in Section~\ref{sec: dataset}. 

    \item In terms of model outputs, there are also three categories to summarize these works: \textbf{scalar}, \textbf{graph}, and \textbf{action sequence}. Scalar outputs are usually used in prediction tasks which provide indicators of the epidemic like infected cases. There are also some works that focus on epidemic graph construction and the outputs of their designed models are graphs ~\cite{shan2023novel, wang2023wdcip}. Finally, the projection tasks we collected usually adopt Reinforcement Learning (RL), which outputs the actions taken at each time step, forming a consecutive action sequence  ~\cite{jhun2021effective, fengContactTracingEpidemic2023, meirom2021controlling, song2020reinforced}.


\end{compactenum}

\subsection{Data Sources}
\label{sec: dataset}

\begin{figure}[t]
	\begin{center}
    \includegraphics[width=0.98\linewidth]{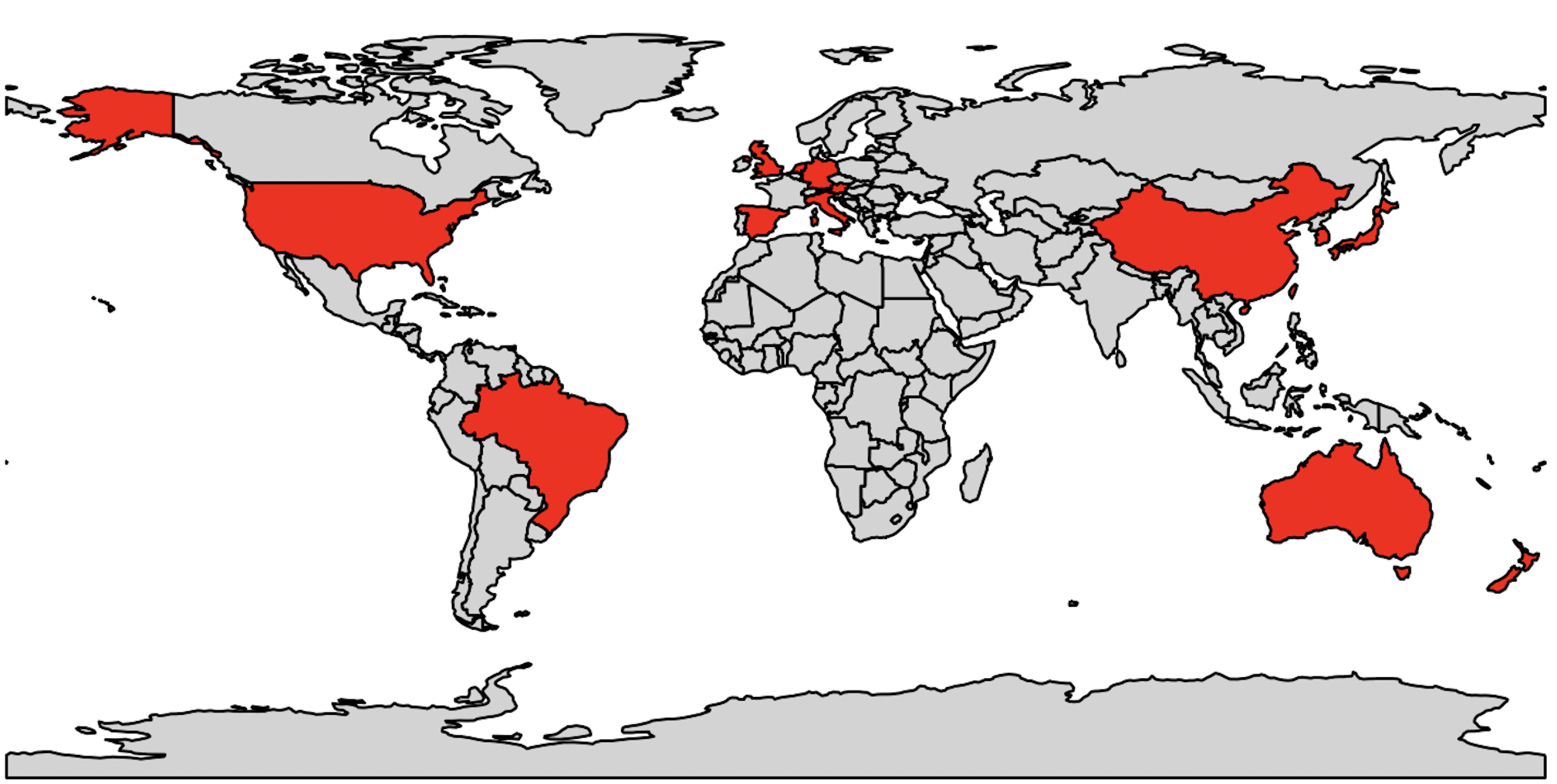}
	\end{center}
	\vspace{-5pt}
	\caption{{Countries involved in sampling epidemiology data.} 
 }
	\vspace{-5pt}
	\label{fig: map}
 \vspace{-10pt}
\end{figure}


The investigated works in this survey cover a wide range of datasets from different parts of the world, as shown in Figure ~\ref{fig: map}. However, the majority of research focuses only on COVID-19 while only a few study Influenza-Like Illnesses~(ILI) or bacteria ~\cite{gouarebDetectionPatientsRisk2023}. We  categorize these datasets based on their sources:
\begin{compactenum}[(1)]
    \item \textbf{Demographic and Health Records.} Epidemic data can be accessed through public databases released by universities, governments, or other organizations. These data usually include demographic information like populations, number of infections, and health records of individuals or groups. During epidemic graph construction, these data are usually used directly as node features or used in the construction of graph structures. 

    \item \textbf{Mobility Information.} Mobility information can be acquired through websites that record transportation information, maps, or contact records of individuals. This information is usually used to construct the graph structure.

    \item \textbf{Online Search and Social Media.} Epidemic information can also be acquired through social media and online search records. The massive search of disease-related questions in a region can indicate potential outbreaks~\cite{lin2023grapha}, which can then be utilized as node features.

    \item \textbf{Sensors.} Multi-modal data can be acquired through sensors like cameras, satellites, radios, etc. These data can also help epidemic tasks like exposure risk prediction using images ~\cite{hanDevilLandscapesInferring2023}. Unlike conventional data sources, sensor data often requires additional preprocessing using specialized models, such as encoding images with techniques like ResNet~\cite{he2016deep}, before integration as node features.

    \item \textbf{Simulated Data.} Besides real-world data, some research ~\cite{tang2023enhancing, sunDeepDynaForecastPhylogeneticinformedGraph2023, meznar2021prediction} also utilized simulated data for model training and testing. These data often require simulation models like TimeGEO ~\cite{jiang2016timegeo}, Independent Contagion Model (ICM) ~\cite{murphy2021deep}, and also SIR models to generate temporal graphs.
  
\end{compactenum}

\subsection{Graph Construction}
\label{sec: Graph_Construct}
For graph construction, we provide a taxonomy based on the dynamicity of nodes and edges as follows.

\subsubsection{Static Node Features.}
Static node features typically refer to characteristics that do not change with time. The shape of static features can be represented as $\mathbb{R}^{N \times h} $, where $h$ refers to the number of different features. Besides tasks involving time series, most GNN tasks are using static features. For example, in a contact graph in which individuals are modeled as nodes and contact information represents edges, personal characteristics like infection order, gender, age, and symptom-related information can be used as static features during training and prediction ~\cite{songCOVID19InfectionInference2023}. However, tasks involving time series can also use static features as additional information.

\subsubsection{Dynamic Node Features.}
Contrary to static features, dynamic features represent characteristics that change through time. This type of data is commonly seen in time-series forecasting tasks and the models usually require inputs at each time point. Therefore, the shape of the dynamic features can be represented as $ \mathbb{R}^{N \times T \times h}$, where $T$ refers to the number of time points given. As an example, the number of daily confirmed cases in a region can be seen as dynamic features ~\cite{xieEpiGNNExploringSpatial2022}. Although most models take in a single slice of dynamic features at each time point, some models use the entire dynamic features across time $T$ in a single input ~\cite{ruInferringPatientZero2023}.

\subsubsection{Static Graph Structure.}
The construction of a static graph structure typically entails the use of a predefined approach to generate the graph from available data. Once the graph is established, its structure remains unchanged throughout the training iterations or over different time points. For instance, in tasks involving multiple regions, the geographical adjacency $A$, is often employed to connect different regions, which are represented as nodes in the graph $G$ ~\cite{pu2023dynamic, yu2023spatiotemporal}. The distance between regions or other features can be considered as the edge weights. Another strategy focuses on exploring human mobility or transitions, \eg{}, linking nodes through nearest neighbors in the case of COVID-19 transmission. This method takes into consideration the distribution of the population and individual movements between various locations ~\cite{liu2023human, wang2023wdcip}. In research where the node represents an individual, connections between two nodes often utilize contact information, \eg{}, identifying contacts at risk of spreading the disease as links between individuals ~\cite{gouareb2023detectiona, tomy2022estimating}.

\subsubsection{Dynamic Graph Structure.}
Determining the structure of a dynamic graph commonly involves one of two methodologies. One approach is the modification of adjacency relations over time or throughout the virus propagation process. For example, ~\cite{meirom2021controlling} utilize $\mathcal{E}(t) = \{e_{uv}(t)\}$ to represent the set of edges at time step $t$, which connect individuals based on transmission probability. Another strategy entails the learning of adaptive edges or edge weights during the training phase. Given the dynamic nature of disease transmission, which evolves at each time step, traditional geographical adjacency matrices fall short of accurately representing true connectivity. Recent studies ~\cite{wangCausalGNNCausalBasedGraph2022, deng2020colagnn, caoMepoGNNMetapopulationEpidemic2022} have aimed for models to learn an adaptive relationship between nodes. This typically involves initially generating node features via a neural network, followed by the computation of an attention matrix to depict dynamic connectivity, often expressed as \( {\bf A}_t = a_{i,j} \in \mathbb{R}^{N \times N} \), where \( a_{i,j} \) indicates the influence of node \( v_j \) on node \( v_i \).

\subsection{Methodological Distinctions}
\label{sec:neural_vs_hybrid}
The methodologies of the GNNs in epidemic modeling can be broadly classified into two categories: \textbf{Neural Models} and \textbf{Hybrid Models}.
This classification illuminates the extent of methods that combine computational techniques with epidemiological insights. Both categories employ neural networks, yet they diverge in their underlying principles. (a) Neural Models primarily focus on a data-driven approach and leverage the power of deep learning (i.e., GNNs in our paper) to uncover complex patterns in disease dynamics without explicit encoding of the underlying epidemiological processes. (b) On the other hand, Hybrid Models represent a synergistic fusion of mechanistic epidemiological models with neural networks.    This integration allows for the structured, theory-informed insights of mechanistic models to complement the flexible, data-driven nature of GNNs, aiming to deliver predictions that are interpretable, accurate, and grounded in theoretical knowledge.



\section{Methodology}
\label{sec: Method}

\begin{figure}[t]
	\begin{center}
    \includegraphics[width=0.98\linewidth]{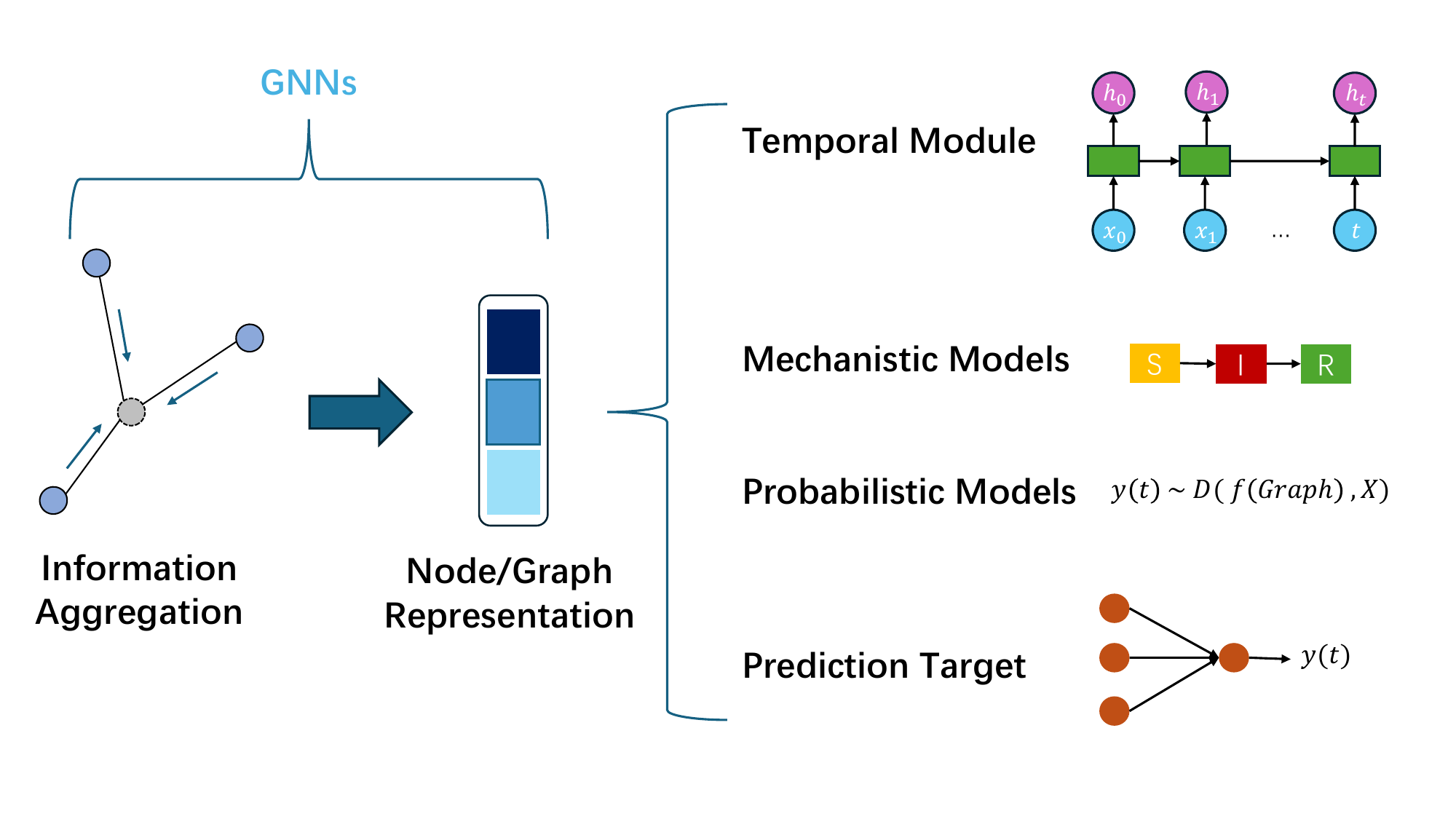}
	\end{center}
	\vspace{-5pt}
	\captionsetup{font=small}
	\vspace{-2em}	
 \caption{GNNs aggregate information from neighborhoods. After aggregation, the node/graph representation can be further utilized in the temporal module, employed to predict parameters of mechanistic and probabilistic models, or directly output prediction targets.
 }
\vspace{-2em}
\label{fig:gnn}
\end{figure}

In this section, we provide a detailed illustration of the methods in epidemic modeling,  which are divided into the two categories discussed in Section~\ref{sec:neural_vs_hybrid}:  \textit{Neural Models} and \textit{Hybrid Models}. While both categories utilize GNNs as the backbone model as illustrated in Figure~\ref{fig:gnn}, they differ in the adoption of the mechanistic models. 

\subsection{Neural Models}

When utilizing GNNs for epidemic modeling, numerous studies have exclusively employed GNNs without incorporating mechanistic models into their tasks, which we term Neural Models. These data-driven models can automatically learn features from raw data and capture intricate patterns across diverse inputs. This inherent capability significantly enhances their performance across various tasks. 
In this subsection, we delve into the (GNN-based) Neural Models in epidemic modeling, dissected through three perspectives: (a) \textbf{Spatial Dynamics Modeling}, (b) \textbf{Temporal Dynamics Modeling}, and (c) \textbf{Intervention Modeling}. 
This categorization is designed to specifically tackle the challenges of modeling the spatial spread, temporal evolution, and the impact of intervention strategies through the advanced capabilities of GNNs.





\subsubsection{Spatial Dynamics Modeling} 
One advantage of GNNs,  e.g., GCN or GAT, is their ability to capture spatial relationships through various aggregation processes, which can analyze and capture the spatial dimensions of disease propagation. Numerous studies represent the inherent structure of geographical data as graph data, denoted as ${\bf A}$, where nodes depict regions (e.g., cities, neighborhoods, or countries), and edges describe connections between these regions (e.g., roads, flights, or potential vectors for disease transmission). 
Subsequently, GNNs are applied to the graph data to uncover complex relationships and dependencies at the regional level, facilitating predictions regarding disease spread across different areas~\cite{song2023covid19, pudynamic, nguyen2023predicting, gouareb2023detectiona, lagatta2021epidemiological}. 

In the context of GNN modeling, the significance of edge weights is paramount, as they encapsulate the intensity and nature of interactions. Within epidemiological studies, these weights are often derived from the mobility or social connectedness between regions ~\cite{lin2023grapha, panagopoulos2021transfer}. For instance, studies such as ~\cite{song2020reinforced, cao2023metapopulation} utilize Origin-Destination (OD) flows to quantify inter-regional mobility, thereby dynamically capturing the intensity of transmission. To further enhance the spatial context of each node within the graph, some research advocates for the implementation of positional encoding techniques ~\cite{wang2022equivariant, bograph}. These techniques are designed to augment the nodes' spatial awareness. For example, \textit{Liu et al.}~\cite{liu2023human} introduced a unique encoding for each location, denoted as \(PE(k)\), with even and odd elements represented by \(\sin(k/10000^{i/L})\) and \(\cos(k/10000^{i-1/L})\) respectively, where \(L\) denotes the dimension of the encoding. 

While GNNs have shown success in modeling spatial relations, challenges arise when dealing with varying input data. Specifically, the absence of direct structural information and the introduction of more complex structural information pose additional difficulties during modeling. To tackle these challenges, several studies have attempted solutions, as outlined below.


\vskip 0.3em
\noindent\textbf{Adaptive Structure Learning.} Although GNNs possess the inherent capability to learn the spatial characteristics of disease dissemination, there are occasions when adjacency relationship information is not available in the real world, often due to data scarcity. To overcome this challenge, several studies highlight the importance of learning an adaptive structure throughout the training process \cite{zheng2021hierst, wang2022adaptively}. For instance, \textit{Wang et al.}~\cite{wang2022adaptively} introduced a graph structure learning module, denoted as \(f_\theta\). This module is designed to calculate node similarities, thereby representing spatial relationships as follows:
\begin{equation}
\begin{gathered}
\label{eq: structure}
\small
\setlength\abovedisplayskip{5pt} \setlength\belowdisplayskip{5pt}
\mathbf{M}_1 = \tanh(f_{\theta_1}(\alpha \mathbf{X}_1)), \quad \mathbf{M}_2 = \tanh(f_{\theta_2}(\alpha \mathbf{X}_2)),  \\
\mathbf{A} = \text{ReLU}(\tanh(\alpha (\mathbf{M}_2 \mathbf{M}_2^\top - \mathbf{M}_1 \mathbf{M}_1^\top))),
\end{gathered}
\end{equation}
where \(\mathbf{X}_1\) and \(\mathbf{X}_2\) are randomly initialized, learnable node embeddings, while \(\alpha\) represents a hyper-parameter. \textit{Shan et al.}~\cite{shan2023novel} employed a method to estimate the graph Laplacian from COVID-19 data through convex optimization of derived eigenvectors. This approach aims to identify dynamic patterns of pandemic spread among countries by analyzing their structural relationships. Additionally, inspired by recent advancements in attention-based mechanisms \cite{vaswani2023attention, turner2024introduction, soydaner2022attention}, a considerable portion of research suggests the use of an attention matrix to illustrate the relationships between nodes \cite{deng2020colagnn, wangCausalGNNCausalBasedGraph2022, cui2021unobservables}. Notably, Cola-GNN \cite{deng2020colagnn} pioneers the application of additive attention in learning the adaptive structure, which is defined as follows:
\begin{equation}
\label{eq: attention}
a_{i,j} = \mathbf{v}^T g(\mathbf{W}^s\mathbf{h}_i + \mathbf{W}^t\mathbf{h}_j + \mathbf{b}^s) + b^v,
\end{equation}
where \(g\) is an activation function, \(\mathbf{W}^s, \mathbf{W}^t \in \mathbb{R}^{d \times D}\), \(\mathbf{b}^s \in \mathbb{R}^{d}\), and \(b^v \in \mathbb{R}\) are trainable parameters, with \(d\) as a hyperparameter controlling the dimensions of these parameters. \(a_{i,j}\) reflects the impact of location \(j\) on location \(i\). This approach allows for dynamic adaptation to changes in graph structure, effectively capturing asymmetric and complex viral transmission patterns. 

\vskip 0.3em
\noindent\textbf{Multi-Scale Modeling.} Previous approaches typically operate at a singular level, overlooking the multifaceted nature of real-world epidemiological data, which encompasses multiple scales such as country, state, and community levels. For epidemiological tasks, multi-scale modeling is imperative for capturing the dynamics of disease spread across these varied levels, from individual behaviors to global dissemination, ensuring a more comprehensive analysis. HierST \cite{zheng2021hierst} leverages multi-scale modeling to effectively capture the spread of COVID-19 across different administrative levels by constructing a unified graph, which encapsulates the spatial correlation dynamics both within and between these levels. To further address both local interactions and long-range dependencies, MSGNN \cite{qiu2023msgnna} is meticulously designed to integrate influences from both immediate and broader regions on disease transmission, enhancing its effectiveness in cross-scale epidemiological dynamics.

\subsubsection{Temporal Dynamics Modeling}
The temporal dynamics in epidemiological models are pivotal for capturing the evolution of disease spread, reflecting changes in infection rates, recovery rates, and other critical parameters over time. These models typically conceptualize graphs as spatio-temporal networks, underscoring the significance of temporal data in comprehending disease dynamics, and forecasting future trends~\cite{zhang2023predicting,sijirani2023spatiotemporala}. A particular strand of research utilizes RNN-based models (\eg{}, LSTM or GRU), as mechanisms to extract node features. These features are then incorporated into the graph convolution process \cite{CoLA_TNNLS21, zheng2021hierst}. A simple way \cite{kapoor2020examining} to achieve this by executing the concat operator:
\begin{equation}
\label{eq: h}
\setlength\abovedisplayskip{5pt} \setlength\belowdisplayskip{5pt}
\textbf{h} = \text{MLP}(\textbf{x}_t|\textbf{x}_{t-1}|...|\textbf{x}_{t-d}) 
\end{equation}
where $\textbf{h}$ is simply the output of an MLP (Multilayer Perceptron) over the node temporal features $\textbf{x}$ at time $t$ reaching back $d$ days.
Another surge of approaches first executes graph spatial convolution in each time step separately, and then leverages all outputs of the GNNs as the input of the temporal module and utilizes them for final downstream tasks like the prediction \cite{gao2021stan, yuSpatiotemporalGraphLearning2023,moon2023graph, duarte2023time}. STEP \cite{yuSpatiotemporalGraphLearning2023} execute the multi-layers graph convolution operation to get all node embedding $\textbf{h}$, and then leverage the GRU to  
get the final output:
\begin{equation}
\label{}
\textbf{h}_t = \textbf{z}_t \circ \textbf{h}_{t-1} + (1 - \textbf{z}_t) \circ \textbf{h}'_t,
\end{equation}
where \( \textbf{h}_t \) is the final result, and \( \textbf{z}_t \) is the result of the update gate, which controls the inflow of information in the form of gating. The Hadamard product of \( \textbf{z}_t \) and \( \textbf{h}_{t-1} \) represents the information retained to the final memory at the previous timestep.

In contrast to the initial two methodologies, numerous studies achieve their final output by iteratively layering GNN and temporal models \cite{sesti2021integrating}. Some work~\cite{guo2021research, sha2021source} advocate for the employment of Spatio-Temporal Graph Neural Networks (STGNNs)~\cite{hu2022graph, zhou2021graph, wu2021comprehensive} to extract insights from multivariate spatiotemporal epidemic graphs. An STGNN integrates many ST-Conv blocks, which comprise a spatial layer flanked by two temporal layers. Each temporal layer features a 1-D CNN operating along the time axis, followed by a Gated Linear Unit (GLU), to delineate the temporal dynamics. The spatial layer, on the other hand, utilizes a GCN based on the Chebyshev polynomials approximation \cite{Cheb_Neurips16, Cheb2_Neurips22} for spatial analysis.
To further refine the understanding of spatial dynamics during disease evolution, RESEAT ~\cite{moon2023reseat} proposes the continuous maintenance and adaptive updating of an attention matrix. This process aims to capture regional interrelationships throughout the entirety of the input data period:
\begin{equation}
\begin{gathered}
\label{}
tp^t_{i,j}(\textbf{A}) = \textbf{A}^t_i \cdot \textbf{A}^t_j, \\ 
\textbf{A}^{t+1}_{i,j} = \text{softmax}(a^{t+1}_{i,j} + tp^t_{i,j}(\textbf{A})), \\
Att_i = \sum_{j=1}^N \textbf{A}^t_{i,j} \times \textbf{x}_j.
\end{gathered}
\end{equation}
The $\textbf{A}^t_{i,j} \in \mathbb{R}$ denotes the attention weight between regions $i$ and $j$ at time step $t$, and $Att_i$ is employed as the final feature for the node $v_i$. Through this mechanism, RESEAT adeptly captures not only temporal patterns but also the dynamically evolving regional interrelationships. To integrate explicit observations with implicit factors over time, \textit{Cui et al.}~\cite{cui2021unobservables} introduced a new case prediction methodology within an encoder-decoder framework. They contend that relying solely on observed case data, which can be inaccurate, may impair prediction performance. Accordingly, their proposed decoder is designed to incorporate inputs of new cases and deaths, thereby dynamically reflecting temporal changes.

\subsubsection{Intervention Modeling}
Intervention modeling offers a detailed perspective on epidemic spread by simulating the behaviors and interactions of individuals within a network based on intervention strategies.
This method provides an intricate view of individual actions, mobility patterns, and the likelihood of disease transmission. When combined with GNN, this approach enhances the model's capability to represent the diversity and complexity inherent in real-world social networks, augmenting the efficacy of intervention strategies. \textit{Song et al.}~\cite{song2020reinforced} introduced a reinforcement learning framework that dynamically optimizes public health interventions to strike a balance between controlling the epidemic and minimizing economic impacts, to reduce infection rates while maintaining economic activities.
To delve deeper into the individual underlying dynamics, \textit{Meirom et al.}~\cite{meirom2021controlling} proposed a dual GNN module strategy. One module updates the node representations according to dynamic processes, while the other manages the propagation of long-range information. Subsequently, they employ RL to modulate the dynamics of social interaction graphs and perform intervention actions on them. This approach aims to indirectly curb epidemic spread by strategically altering network structures, thus avoiding direct intervention in the disease process.

IDRLECA \cite{feng2023contact} embodies a novel integration, combining an infection probability model with an innovative GNN design. The infection probability model calculates the current likelihood of each individual's infection status. This information, along with personal health and movement data, is utilized to forecast virus transmission through human contacts using the GNN:
\begin{equation}
\begin{gathered}
\label{}
p_{i,\text{infected}} = 1 - \hat{p}_{i,\text{healthy}} = 1 - p_{i,\text{healthy},T} \times (1 - p_c)^{\text{contacts}}, 
\end{gathered}
\end{equation}
here \(p_{i,\text{infected}}\) represents the probability that individual \(i\) is infected, while \(p_{i,\text{healthy}, T}\) denotes the baseline probability of individual \(i\) being healthy at time \(T\), before accounting for contact-related risks. \(p_c\) refers to the probability of infection from a single contact. Additionally, a custom reward function is designed to simultaneously minimize the spread of infections and the associated costs, striking a balance between health objectives and economic considerations:
\begin{equation}
\begin{gathered}
\label{}
   r = -\left( \exp\left(\frac{\Delta I}{\theta_I}\right) + \exp\left(\frac{\Delta Q}{\theta_Q}\right) \right).
\end{gathered}
\end{equation}
This function considers the change in the number of infections (\(\Delta I\)) and the cost of mobility interventions (\(\Delta Q\)), with \(\theta_I\) and \(\theta_Q\) acting as soft thresholds for these changes.

\subsection{Hybrid Models}
Unlike Neural Models described above, Hybrid Models effectively combine the predictive capabilities of neural networks with the foundational principles of mechanistic models, thereby enhancing both the accuracy and interpretability of disease forecasting. This integration can be further classified into two categories: \textit{Parameter Estimation for Mechanistic Model} and \textit{Mechanistic Informed Neural Model}. The former approach allows these hybrid systems to adapt to evolving epidemic patterns by dynamically estimating parameters of mechanistic models using neural networks, ensuring that simulations remain closely aligned with current trends. Conversely, the latter approach involves incorporating priors from mechanistic models into neural networks, enriching these models with domain-specific knowledge, and directing the learning process to more accurately reflect plausible disease dynamics. This synergistic methodology not only capitalizes on the data-driven strengths of neural models but also firmly anchors predictions within the framework of epidemiological theory, presenting a comprehensive and informed strategy for predicting epidemic spread.

\subsubsection{Parameter Estimation for Mechanistic Models}
This line of research highlights that hybrid models, which integrate neural networks, dynamically adjust the parameters of mechanistic models. This combination enables the analysis of real-time data, thus informing and refining mechanistic models to ensure their simulations accurately mirror the dynamics of actual epidemics \cite{jhun2021effective, tang2023enhancing, lagatta2021epidemiological}. Notably, studies like \cite{lagatta2021epidemiological, xie2022visualization} employ GNNs to estimate the contact (transmission) rate, $\beta$, and to monitor the epidemic evolution through the implementation of the SIR model. Further, research~\cite{gao2021stan, tang2023enhancing} estimates both the transmission rates $\beta$ and recovery rates $\gamma$ by leveraging outputs from the GNN. This methodology initiates with the utilization of GRU to derive node embeddings $\textbf{h}$, which subsequently facilitate the calculation of parameters:
\begin{equation}
\label{}
\beta,\gamma = \text{MLP}_1(\textbf{h})\quad\quad
\Delta I, \Delta R = \text{MLP}_2(\textbf{h}),
\end{equation}
where $\Delta I$ and $\Delta R$ denote the daily increases in the number of infected and recovered cases, respectively. To enhance the model's ability to leverage the dynamics of the pandemic for regulating longer-term progressions, the researchers utilize the predicted transmission and recovery rates to calculate predictions based on the dynamics of the disease spread:
\begin{equation}
\begin{gathered}
\label{}
\hat{\Delta I}^d = \left[ \hat{\Delta I}^d_{t+1}, \hat{\Delta I}^d_{t+2}, ..., \hat{ \Delta I}^d_{t+L_p} \right], \\
\text{ each } \hat{ \Delta I}^d_i = \beta S_{i-1} - \gamma I_{i-1} = \beta(N_p - \hat{I}^d_{i-1} - \hat{R}^d_{i-1}) - \gamma \hat{I}^d_{i-1}, \\
\hat{\Delta R}^d = \left[ \hat{ \Delta R}^d_{t+1}, \hat{\Delta R}^d_{t+2}, ..., \hat{\Delta R}^d_{t+L_p} \right], \text{ each } \hat{\Delta R}^d_i = \gamma I^d_{i-1}, \\
\end{gathered}
\end{equation}
where $I^d_{i-1}$ and $R^d_{i-1}$ are iteratively calculated using the ground truth of the infected and recovered cases from the day preceding the current prediction window. $N_p$ represents the population size of the current location, $t$ denotes the time steps, and $L_p$ refers to the number of days into the future for which predictions are made. Ultimately, the researchers propose two loss functions to consider both the short-term and long-term progression of the pandemic.

To go beyond single-region recognition, MepoGNN \cite{cao2023metapopulation, cao2023mepognn} extends the SIR model to the metapopulation variant \cite{Data-centric_arxiv22, balcan2009multiscale, mousavi2012enhanced}, accommodating heterogeneity within populations and incorporating human mobility to model the spread between sub-populations: 
\begin{equation}
\begin{gathered}
\label{}
\frac{dS_i(t)}{dt} = -\beta_i(t) \cdot S_i(t) \sum_{j=1}^N \left( \frac{h_{ji}(t)}{P_j} + \frac{h_{ij}(t)}{P_i} \right) I_j(t), \\
\frac{dI_i(t)}{dt} = \beta_i(t) \cdot S_i(t) \sum_{j=1}^N \left( \frac{h_{ji}(t)}{P_j} + \frac{h_{ij}(t)}{P_i} \right) I_j(t) - \gamma_i(t) \cdot I_i(t), \\
\frac{dR_i(t)}{dt} = \gamma_i(t) \cdot I_i(t).
\end{gathered}
\end{equation}
MepoGNN incorporates a spatio-temporal GNN designed to learn three dynamic parameters: $\beta_i(t+1)$, $\gamma_i(t+1)$, and $\textbf{H}(t)$, throughout the evolving timeframe. Here, $\textbf{H}(t)$ signifies the epidemic propagation matrix, capturing human mobility between regions, represented by $\{h(t)_{ij}|i, j \in \{1,2,..., N\}\}$. The model thereby generates its final prediction of daily confirmed cases as follows:
\begin{equation}
\begin{gathered}
\label{}
y_i(t) = \beta_i(t) \sum_{j=1}^N \left( \frac{h_{ji}(t)}{P_j} + \frac{h_{ij}(t)}{P_i} \right) I_{j}(t), 
\end{gathered}
\end{equation}
Recent work \cite{liu2023epidemiologyaware} integrates the Cola-GNN \cite{deng2020colagnn} framework with the SIR model through the development of Epi-Cola-GNN, introducing a mobility matrix $\Pi$ to capture the dynamics of infectious disease spread across different locations. Within this matrix, $\pi_{ij}$ quantifies the intensity of human mobility from location $i$ to location $j$, offering a nuanced perspective on the spatial transmission of diseases. This incorporation leads to a modification in the representation of infectious cases within the SIR model framework:
\begin{equation}
\begin{gathered}
\label{}
\frac{dI_i}{dt} = \beta_i I_i - \gamma_i I_i - \sum_{j=1, j \neq i}^N \pi_{i,j} I_i + \sum_{j=1, j \neq i}^N \pi_{j,i} I_j.
\end{gathered}
\end{equation}
Furthermore, they introduce the concept of the Next-Generation Matrix (NGM) \cite{diekmann2010construction}, which provides a clearer epidemiological interpretation and more effectively supports both intra-location spread and inter-location transmission influenced by human mobility. 

Instead of simply estimating the rate indicator, EpiGCN \cite{han2023devila} innovatively employs three distinct linear layers to transform the node feature into the SIR state, enhancing the model awareness of the available data:
\begin{equation}
\begin{gathered}
\label{}
S_v = \sigma(\textbf{W}_s \cdot \textbf{h}_v + b_s),  I_v = \sigma(\textbf{W}_i \cdot \textbf{h}_v + b_i),  R_v = \sigma(\textbf{W}_r \cdot \textbf{h}_v + b_r). 
\end{gathered}
\end{equation}
Subsequently, they refine the process of updating the SIR \cref{eq: SIR} model and introduce a novel SIR message-passing mechanism that aggregates information from neighboring nodes. This approach modifies the conventional SIR update equation to incorporate spatial dependencies and interactions within the network:
\begin{equation}
\begin{gathered}
\label{}
S_v = S_v - \textbf{W}_{tran} \cdot \text{concat}\left(S_v, \sum_{\textbf{w} \in \textbf{A}_v} e_w I_w\right), \\
I_v = I_v + \textbf{W}_{tran} \cdot \text{concat}\left(S_v, \sum_{\textbf{w} \in \textbf{A}_v} e_w I_w\right) - \textbf{W}_{recov} I_v, \\
R_v = R_v + \textbf{W}_{recov} \cdot I_v, 
\end{gathered}
\end{equation}
here $\textbf{W}_{tran} \in \mathbb{R}^{2D \times D}$ and $\textbf{W}_{recov} \in \mathbb{R}^{D \times D}$ denote the matrices for linear transformations corresponding to the transmission and recovery processes, respectively. Ultimately, the SIR states are concatenated and transformed to align with the prediction objectives:
\begin{equation}
\begin{gathered}
\label{}
y_v = \text{softmax}\left(\textbf{W}_{output} \cdot \text{concat}(S_v, I_v, R_v)\right).
\end{gathered}
\end{equation}

\subsubsection{Mechanistic-Informed Neural Models}
Unlike previous methods wherein neural networks dynamically adjust the parameters of mechanistic models based on data inputs, \textit{mechanistic-informed neural models} utilize domain knowledge from mechanistic models to inform the architecture and learning processes of GNNs. This strategy flexibility allows for a swift adaptation to changing conditions, markedly improving the accuracy of forecasts and the effectiveness of interventions. Certain studies \cite{meznar2021prediction, wang2023wdcip, tomy2022estimating} utilize the SIR model to generate target data by simulating epidemic spreads from individual nodes, which are then employed to train GNNs for downstream tasks. In the context of source detection tasks, such as those discussed in \cite{ru2023inferring, shah2020finding, sha2021source}, one-hot encoded node states \( x^t_i \in \{0, 1\}^M \), with \( M \) representing the number of possible states, are used as inputs for the GNN, where the states are defined as either \{S, E, I, R\} or \{S, I, R\}. 
\textit{Song et al.}~\cite{song2020reinforced} integrated SIHR (a variant of SIR)~\cite{kermack1927contribution} simulation environment with the RL framework, providing a dynamic model of epidemic progression for the RL agent. This capability allows the agent to account for individuals who are hospitalized, enabling the dynamic modification of mobility control policies.

To explicitly capture causal dynamics, CausalGNN~\cite{wang2022causalgnn} introduces a novel approach to causal modeling by leveraging causal features \( \textbf{Q}_t = (q_{i,t}) \in \mathbb{R}^{N \times 4} \), with \( q_{i,t}: S_i(t), I_i(t), R_i(t), D_i(t) \) representing the cumulative number of individuals in each state of the SIRD model. A causal encoder is then designed to transform these causal features into node embeddings, operating as follows:
\[
\textbf{H}^t_c = \tanh(\textbf{Q}_t\textbf{W}^t_e + b^t_e) \in \mathbb{R}^{N \times D},
\]
where \( \textbf{W}^t_e \in \mathbb{R}^{4 \times D} \) and \( b^t_e \in \mathbb{R}^{D} \) denote model parameters, and these causal features are intended to be concatenated with other node embeddings. The spatial GNN architecture also infers the SIRD rates \( \beta_i(t), \gamma_i(t), \rho_i(t) \) by providing \( \textbf{P}_t = (p_{i,t}) \in \mathbb{R}^{N \times 3} \), which are subsequently utilized for SIRD model updates.

\section{Future Work}
\label{sec: Future_Work}

While many challenges have been addressed in the application of GNNs within epidemic modeling, this field continues to confront various difficulties, both explored and unexplored. In this section, we will examine these challenges and highlight potential avenues for future research.


\subsection{Epidemic at Scales}


Multi-scale data are crucial in epidemiology because they offer comprehensive insights into both intra-region and inter-region relationships, thus aiding in the more accurate modeling of disease spread. Presently, several studies have acknowledged this importance and initiated the integration of multi-scale data into their frameworks ~\cite{tangEnhancingSpatialSpread2023, wangWDCIPSpatiotemporalAIdriven2023, qiuMSGNNMultiscaleSpatiotemporal2023, zheng2021hierst}. Although these efforts have yielded models capable of accommodating multi-scale data, existing approaches are limited to processing only two predefined scales, such as county-level and state-level data. Looking ahead, there is growing anticipation for the development of novel models capable of incorporating data across multiple dynamic scales and adaptable to diverse epidemiological tasks.

Meanwhile, scalability must also be considered for numerous reasons: (1) A smaller granularity results in the expansion of graph data. (2) Some tasks require real-time processing ~\cite{wangWDCIPSpatiotemporalAIdriven2023}. While the number of countries or provinces can be small, the graph for individuals or other necessary parts in epidemic models can be extremely large, e.g., contact information graphs in metropolises, which could make the current methods very time-consuming. Furthermore, the use of multi-scale data and the requirements for real-time processing make the problem even harder.

\subsection{Cross-Modality in Epidemiology}
The integration of multi-modal data in epidemiological tasks offers a powerful approach for enhancing our understanding of disease transmission dynamics, improving predictive accuracy, enabling early detection and intervention, conducting comprehensive risk assessments, and fostering interdisciplinary collaboration to address public health challenges more effectively. Data from different modalities can not only serve as augmentations for each other but also compensate for noise from single-modality data. In recent years, some works have successfully incorporated multi-modality in GNNs. Although GNNs are very suitable for information aggregation and handling multi-modality data, there has not been much work exploring the multi-modality of GNNs in an epidemiology setting. Some related works ~\cite{linGraphNeuralNetwork2023, hanDevilLandscapesInferring2023} have utilized unstructured data like textual or image data to construct node features. However, there is no cross-modality in their work in terms of node features.

\subsection{Epidemic Diffusion Process}
The diffusion process, which is the key component in epidemiological tasks, can be both spatial and temporal. All GNN-based methods discussed above involve information aggregation at one or several time points in a discrete manner. Nevertheless, in the real world, disease spreading is a continuous process, which is incompatible with current methods. To address this problem, Continuous GNNs ~\cite{xhonneux2020continuous, chamberlain2021grand, poli2019graph}, inspired by Neural ODE ~\cite{chen2018neural}, can be applied to model the continuous spreading process.

Another problem lies in that both disease spreading and infection take time, and they can happen asynchronously. One related work ~\cite{puDynamicAdaptiveSpatio2023} considers different time-space effects and models the effects using the attention mechanism. However, it is still done in a discrete manner, creating gaps in the real-world transmission process.

\subsection{Interventions for Epidemics}
In epidemiology, control measures are vital for controlling disease spread and safeguarding public health. They include intervention strategies like vaccination, quarantine, and public health education to limit transmission and minimize outbreaks, ultimately saving lives and reducing the burden on healthcare systems ~\cite{bhattacharyya2023public,jamison2007disease,luo2021what}.
Among the methods mentioned in this paper, most of the research incorporates intervention strategy in agent-based models ~\cite{fengContactTracingEpidemic2023, meirom2021controlling, song2020reinforced} or other neural models ~\cite{pu2023dynamic}. Generally, the interventions in these methods include deleting nodes, altering nodes, and altering edge weights. However, each method only includes one type of intervention, either node-level or edge-level. In the real world, however, interventions can happen at different graph levels and also at different scales. To better model the real situation, multi-level and multi-scale interventions need to be introduced.

\subsection{Generating Explainable Predictions}

The study of epidemic modeling not only aims for accurate predictions but also emphasizes interpretability. Ideally, experts will rely on epidemic models' predictions to make informed decisions. However, relying on a model's predictions becomes risky if the model cannot provide confidence in its forecasts, given the significant consequences of these decisions. Therefore, interpretability is essential and offers various benefits, including understanding disease dynamics, identifying risk factors, and providing measures of uncertainty. Despite the crucial role of interpretability, neural models investigated thus far have not placed significant emphasis on this aspect, with hybrid models primarily relying on mechanistic models to provide explanations. In recent years, there have been some approaches aimed at providing interpretability for general Graph Neural Networks~\cite{dai2021towards, ying2019gnnexplainer, yuan2022explainability,nian2024globally}, which also hold potential usefulness in epidemiological settings.


\subsection{Handling Challenges from Epidemic Data}
The idea of Data-Centric AI (DCAI) has grown more and more important in recent years ~\cite{zha2023data}, which inspired people to pay attention not only to modeling but also to processing data itself. In epidemiological tasks, there are also many challenges originating from data, e.g., noise, incomplete data, privacy, etc. Although there have not been many works addressing these challenges in GNN-related epidemiological tasks, we expect future works will try to tackle this problem by proposing model-centric and data-centric methods.

\noindent
\textbf{Noisy Data}. 
 \textit{Rodríguez et al.} ~\cite{rodriguezDataCentricEpidemicForecasting2022} introduced several sources of data in epidemiology and noise naturally exists in these data. For example, while social media can provide information on epidemic progress, it may also create significant noise by spreading rumors and misinformation. For GNNs in epidemiology, noise can exist both at node-level and edge-level. So far, the denoising mechanisms for GNNs in epidemiology have remained to be studied. Fortunately, there has been a wide range of studies on the robustness of GNNs~\cite{jin2020graph, jin2021adversarial,zhu2021deep,liu2021elastic}, which may be adapted to epidemic data.

\noindent
\textbf{Incomplete Data}. In addition to noises, epidemic data can also be incomplete. The data-gathering process is not always perfect and can not guarantee the accurate collection of features for every node. This problem may be mitigated during modeling because the blank features can be replaced by the aggregated neighbor features ~\cite{tomy2022estimating} and simulation or interpolation can also be used to infer missing features at some time points. Nevertheless, there is not much work studying the influence of incomplete data in an epidemiology setting while using GNN models. 

\noindent
\textbf{Privacy Protection}. In the real world, epidemic data typically includes sensitive information such as individual health status, location, and potentially identifiable details, which, if mishandled, can lead to severe privacy violations and undermine public trust in health systems ~\cite{selgelid2016ethics, caals2017ethics, liuyang2022interpretationa}. In contemporary research, methods commonly utilize all available epidemic data that span countries or regions. However, this reliance on large-scale data may heighten awareness among government entities and departments regarding data privacy. Owing to the growing focus on data sensitivity, stringent legislations ~\cite{IPR_06, ToFairnessinDataSharing_NEJM16} have been introduced to regulate data collection and utilization. Therefore, future work should take more privacy problems into consideration.
Federated Graph Learning (FGL) ~\cite{FedGCN_arxiv22, FCCL_CVPR22, FLSurveyandBenchmarkforGenRobFair_arXiv23} emerges as a potent solution to these privacy concerns by leveraging the distributed nature of data without necessitating its central aggregation. This approach aligns with the stringent requirements of data privacy regulations by enabling data to remain at its source, thereby minimizing the risks associated with centralized data storage and processing ~\cite{fgssl_IJCAI23, wan2024federated}. 

\section{Conclusion}
In this survey, we present a comprehensive overview of graph neural networks in epidemic modeling. First, we provide introductions and definitions not only for epidemiology and epidemic modeling but also for Graph Neural Networks (GNNs). Then, clear and structured taxonomies for epidemiological tasks and methodology are proposed. In terms of epidemiological tasks, we present a categorization of tasks consisting of four parts: Detection, Surveillance, Prediction, and Projection. In terms of methodology, we focus on GNN-based methods and separate them into Neural Models and Hybrid Models. At the end of our survey, we not only point out the challenges and drawbacks of the current methodology but also offer a number of promising directions for interested researchers to work on. The aim of this survey is to bridge the gaps between Graph Neural Networks (GNNs) and epidemiology, inspiring both epidemiologists and data scientists to pursue advancements in this burgeoning field.

\printbibliography

\clearpage
\onecolumn
\appendix
\appendixpage
\setcounter{table}{0}
\section*{Complete Taxonomy}

\begin{table*}[h]
\caption{
\small{\textbf{Summary of epidemiological tasks and Methodology}.}
}
\label{app: 1}
\vspace{-10pt}
\centering
{
\resizebox{\textwidth}{!}{
\setlength\tabcolsep{10pt}
\renewcommand\arraystretch{1.0}

\begin{tabular}{@{}c|c|c|c|c@{}} 
\hline
\thickhline

\textbf{Task}                          & \textbf{Paper} & \textbf{Methodology} & \textbf{Hybrid}  & \textbf{Graph Construction} \\ 
\hline \hline
\multirow{3}{*}{\textbf{Detection}}    
                              &   SD-STGCN ~\cite{sha2021source}    & GAT + GRU + SEIR &  \hlg{\tmark{}}    &  Spatial-Temporal Graph; Static Graph Structure     \\ \cline{5-5}
                              
                              &   ~\cite{ruInferringPatientZero2023}    & GCN + SIR &  \hlg{\tmark{}} &  \multirow{2}{*}{Spatial-Temporal Graph; Dynamic Graph Structure}     \\
                              &   ~\cite{shah2020finding}    & GNN + SIR    &  \hlg{\tmark{}}  &        \\
                              
\cline{1-5}
\multirow{4}{*}{\textbf{Surveillance}} 
                              &   WDCIP ~\cite{wang2023wdcip}    &  GAE &  &  \multirow{5}{*}{Spatial Graph; Static Graph Structure}          \\
                              &   ~\cite{songCOVID19InfectionInference2023}   & GAT/GCN &     &          \\ 
                              &   ~\cite{gouarebDetectionPatientsRisk2023}    & GCN  &        &        \\
                              &   ~\cite{hanDevilLandscapesInferring2023}    & GCN + Modified SIR &  \hlg{\tmark{}}   &    \\
                              &   ~\cite{shan2023novel}    & Graph Fourier Transform &      &     \\
\cline{1-5}
\multirow{4}{*}{\textbf{Projection}}   
                              &   MMCA-GNNA~\cite{jhun2021effective}    & GNN + SIR + RL &  \hlg{\tmark{}}  &  \multirow{2}{*}{Spatial-Temporal Graph; Static Graph Structure}   \\ 
                              &   DURLECA ~\cite{song2020reinforced}    &  GNN + RL  &     &       \\\cline{5-5}
                              
                              &   IDRLECA ~\cite{fengContactTracingEpidemic2023}    & GNN + RL &    &   \multirow{2}{*}{Spatial-Temporal Graph; Dynamic Graph Structure}               \\
                              &   ~\cite{meirom2021controlling}    & GNN + RL &     &    \\
                              
\cline{1-5}
\multirow{37}{*}{\textbf{Prediction}}
                              &    DGDI ~\cite{liu2023human}   & GCN + Self-Attention &      &  \multirow{2}{*}{Spatial Graph; Static Graph Structure}       \\ 
                              &    ~\cite{sunDeepDynaForecastPhylogeneticinformedGraph2023}   & GNN + LSTM  &      &    \\ \cline{5-5}
                              
                              &    DVGSN ~\cite{zhang2023predicting}   & GNN &    & Temporal-Only Graph; Static Graph Structure              \\ \cline{5-5}
                              
                              &   STAN ~\cite{gao2021stan} & GAT + GRU &       &    \multirow{11}{*}{Spatial-Temporal Graph; Static Graph Structure}    \\
                              &    MSDNet ~\cite{tang2023enhancing}   &  GAT + GRU + SIS  &    \hlg{\tmark{}}   &     \\
                              &    SMPNN ~\cite{lin2023grapha}   &  MPNN + Autoregression &   \hlg{\tmark{}}    &        \\    
                              &    ATMGNN ~\cite{nguyen2023predicting}   & MPNN/MGNN + LSTM/Transformer &      &      \\
                              &    DASTGN ~\cite{puDynamicAdaptiveSpatio2023}   &   GNN + Attention + GRU   &       &       \\
                              &    MSGNN ~\cite{qiu2023msgnna}   &  GCN + N-Beats  &               \\
                              &    STEP ~\cite{yuSpatiotemporalGraphLearning2023}   &  GCN + Attention + GRU  &       &       \\
                              &    ~\cite{panagopoulos2021transfer}   & GNN + LSTM  &       &      \\
                              &    ~\cite{meznar2021prediction}   & Network Centrality + XGBoost  &     &       \\
                              &    ~\cite{tomy2022estimating}   &   GNN + SEIRD    &    \hlg{\tmark{}}    &      \\
                              &    ~\cite{lagatta2021epidemiological} & GNN + LSTM + SIRD &  \hlg{\tmark{}}     &     \\ \cline{5-5}

                              &    GSRNN ~\cite{li2019study}   & GNN + RNN &      & \multirow{9}{*}{Spatial-Temporal Graph; Static Graph Structure}           \\
                              &    ~\cite{fritzCombiningGraphNeural2022} & GNN + SDDR & \hlg{\tmark{}}  &    \\
                              &    ~\cite{xie2022visualization} & GCN + Modified SIR &  \hlg{\tmark{}}   &     \\
                              &    ~\cite{moonGraphBasedDeep2023}   & K-GNN + GRU &    &  \\
                              &    ~\cite{sesti2021integrating}   &  GNN + LSTM  &    &    \\
                              &    ~\cite{davahli2021predicting}   &  GNN + LSTM  &     &    \\
                              &    ~\cite{murphy2021deep}   &   GNN &    &    \\
                              &    ~\cite{kapoor2020examining}   &  GNN + MLP  &    &   \\
                              &    ~\cite{croftForecastingInfectionsSpatiotemporal2023}   &  GAT + GRU  &      &  \\   \cline{5-5}

                              &   Mepo GNN~\cite{cao2023metapopulation, caoMepoGNNMetapopulationEpidemic2022} & (TCN + GCN) + Modified SIR &   \hlg{\tmark{}}   &   \multirow{14}{*}{Spatial-Temporal Graph; Dynamic Graph Structure}  \\
                              &   Epi-Cola-GNN ~\cite{liu2023epidemiologyaware} &  Cola-GNN + Modified SIR  &  \hlg{\tmark{}}     &   \\
                              &   HiSTGNN ~\cite{ma2022hierarchical} &  Hierarchical GNN + Transformer  &    \\
                              &    CausalGNN ~\cite{wang2022causalgnn} & GNN + SIRD &   \hlg{\tmark{}}  &    \\
                              &    ATGCN ~\cite{wang2022adaptively}   &   GNN + LSTM  &       &    \\
                              &    HierST ~\cite{zheng2021hierst}   & GNN + LSTM  &       &    \\
                              &    RESEAT ~\cite{moon2023reseat}   & GNN + Self-Attention &       &     \\
                              &    SAIFlu-Net ~\cite{jungSelfAttentionBasedDeepLearning2022}   & GNN + LSTM &        &      \\
                              &    Epi-GNN ~\cite{xieEpiGNNExploringSpatial2022}   & GCN + Attention + RNN &       &      \\
                              &    Cola-GNN ~\cite{deng2020colagnn}   & GCN + Attention + RNN &     &       \\
                              &    ~\cite{cui2021unobservables}   & GNN  + Attention + LSTM  &      &       \\
                              &    ~\cite{mahmud2021human}   &  GNN + RNN  &       &     \\
                              &    ~\cite{guo2021research}   &  STGCN &        &     \\
                              &    ~\cite{duarte2023time}   & GCRN/GCLSTM &        &      \\

\cline{1-5}
\hline
\end{tabular}}}
\vspace{-10pt}
\end{table*}


\end{document}